\pdfoutput=1
\documentclass[11pt,a4paper,usenames,dvipsnames]{article}
\usepackage{acl}
\usepackage{times}
\usepackage{latexsym}

\usepackage{microtype}

\usepackage[T1]{fontenc}
\usepackage[utf8]{inputenc}

\usepackage{url}
\usepackage{graphicx}
\graphicspath{ {images/} }
\usepackage{amsfonts}
\usepackage{amsmath}
\usepackage{multirow}
\usepackage{makecell}
\usepackage{censor}
\usepackage{microtype}
\usepackage{amsmath,amsfonts,amssymb}
\usepackage{tabularx}
\usepackage{float}
\usepackage[ruled]{algorithm2e}
\usepackage{setspace}
\usepackage{booktabs}

\def\Snospace~{\S{}}

\usepackage{makecell}

\usepackage{tikz}

\definecolor{fl}{HTML}{00C8FF}

\title{An analysis of document graph construction methods for AMR summarization}

\author{Fei-Tzin Lee, Chris Kedzie, Nakul Verma, Kathleen McKeown \\
        Computer Science Department \\ Columbia University \\ \texttt{\{feitzin,kedzie,verma,kathy\}@cs.columbia.edu}}

\begin{document}
\maketitle

\begin{abstract}
    Abstract Meaning Representation (AMR) is a graph-based semantic representation for sentences, composed of collections of concepts linked by semantic relations. AMR-based approaches have found success in a variety of applications, but a challenge to using it in tasks that require document-level context is that it only represents individual sentences. Prior work in AMR-based summarization has automatically merged the individual sentence graphs into a document graph, but the method of merging and its effects on summary content selection have not been independently evaluated. In this paper, we present a novel dataset consisting of human-annotated alignments between the nodes of paired documents and summaries which may be used to evaluate (1) merge strategies; and (2) the performance of content selection methods over nodes of a merged or unmerged AMR graph. We apply these two forms of evaluation to prior work as well as 
   % CK removed these three words to remove a singleton line.
    %to 
    a new method 
    %we present 
    for node merging and show that our new method has significantly better performance than prior work.
\end{abstract}
\section{Introduction}
Abstract Meaning Representation (AMR) is a graph-based semantic representation that aims to capture semantics at a higher level of abstraction than plain text  \citep{banarescu-etal-2013-abstract}. It has been used in a broad variety of applications ranging from information extraction to natural language generation. As originally designed, AMR is intended to represent the meaning of a single sentence, and in tasks such as entailment or paraphrase detection, the context of a single sentence suffices to perform the task.  In tasks requiring longer contexts, however, the sentence-level limitation forces any application of AMR either to consider sentences independently or to link them together in some way. We specifically focus on the use of document-level AMR as an intermediate representation for automatic summarization. In contrast to end-to-end neural approaches to abstractive summarization (e.g., ~\citet{nallapati-etal-2016-abstractive,see-etal-2017-get,chen-bansal-2018-fast,lewis-etal-2020-bart}), the use of an intermediate semantic representation holds promise for more precisely controlling the selected summary content.

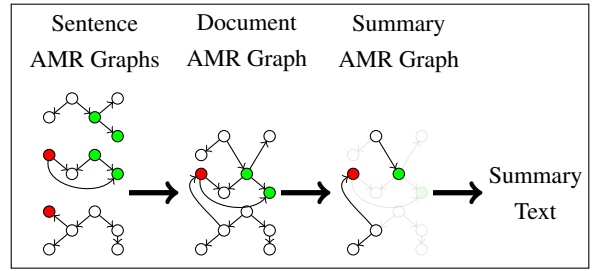
\begin{figure}
\begin{center}
\begin{tikzpicture}[scale=1.0,
amr/.style={draw,circle,minimum width=1.5mm,inner sep=0pt, outer sep=0pt},
]
\draw (-3.8,3.5) rectangle (3.8,0);
\def\sAx{-3.0}
\def\sAy{2.0}
\def\sBx{-3.0}
\def\sBy{1.25}
\def\sCx{-3.0}
\def\sCy{0.5}
\def\Dx{-1}
\def\Dy{1}
\def\Ex{1.0}
\def\Ey{1}
\def\matchcolorA{green}
\def\matchcolorB{red}
\def\skipopacity{0.1}

\node[anchor=north,align=center] at (-3.0+.3,3.5) {\small Sentence\\\small AMR Graphs};

\node[anchor=north,align=center] at (-1+.3,3.5) {\small Document\\\small AMR Graph};

\node[anchor=north,align=center] at (1+.3,3.5) {\small Summary \\\small AMR Graph};

\node[align=center] at (3.1,1) {\small Summary \\\small Text};

\draw[->,line width=0.7mm] (-2.25,1) -- (-1.6,1);
\draw[->,line width=0.7mm] (-2.25+2,1) -- (-1.6+2,1);
\draw[->,line width=0.7mm] (-2.25+4,1) -- (-1.6+4,1);

\node[amr] (nA1) at (\sAx-0.3, \sAy) {};
\node[amr] (nA2) at (\sAx, \sAy + 0.25) {};
\node[amr,fill=\matchcolorA] (nA3) at (\sAx+0.3,\sAy) {};
\node[amr] (nA4) at (\sAx+0.6, \sAy + 0.25) {};
\node[amr,fill=\matchcolorA] (nA5) at (\sAx+0.6, \sAy - 0.25) {};
\draw[->] (nA2) -- (nA1);
\draw[->] (nA2) -- (nA3);
\draw[->] (nA3) -- (nA4);
\draw[->] (nA3) -- (nA5);

\node[amr,fill=\matchcolorB] (nB1) at (\sBx-0.3, \sBy + 0.25) {};
\node[amr] (nB2) at (\sBx, \sBy) {};
\node[amr,fill=\matchcolorA] (nB3) at (\sBx+0.3,\sBy + 0.25) {};
\node[amr,fill=\matchcolorA] (nB4) at (\sBx+0.6, \sBy) {};
 \draw[->] (nB1) -- (nB2);
\draw[->] (nB3) -- (nB2);
\draw[->] (nB3) -- (nB4);
\draw[->] (nB1) to [out=260,in=230] (nB4);

\node[amr,fill=\matchcolorB] (nC1) at (\sCx-0.3, \sCy + 0.25) {};
\node[amr] (nC2) at (\sCx, \sCy) {};
\node[amr] (nC3) at (\sCx+0.3,\sCy + 0.25) {};
\node[amr] (nC4) at (\sCx-0.3, \sCy - 0.25) {};
\node[amr] (nC5) at (\sCx+0.6,\sCy) {};
\node[amr] (nC6) at (\sCx+0.6,\sCy-0.25) {};

\draw[->] (nC2) -- (nC1);
\draw[->] (nC2) -- (nC4);
\draw[->] (nC3) -- (nC2);
\draw[->] (nC3) -- (nC5);
\draw[->] (nC5) -- (nC6);

% \draw[->] (nB1) to [out=260,in=230] (nB4);

\node[amr,fill=\matchcolorB] (nB1) at (\Dx-0.3, \Dy + 0.25) {};
\node[amr] (nB2) at (\Dx, \Dy) {};
\node[amr,fill=\matchcolorA] (nB3) at (\Dx+0.3,\Dy + 0.25) {};
\node[amr,fill=\matchcolorA] (nB4) at (\Dx+0.6, \Dy) {};
 
\node[amr] (nA1) at (\Dx-0.3, \Dy+0.5) {};
\node[amr] (nA2) at (\Dx, \Dy + 0.75) {};
\node[amr] (nA4) at (\Dx+0.6, \Dy + 0.75) {};
\node[amr] (nC2) at (\Dx, \Dy-0.5) {};
\node[amr] (nC3) at (\Dx+0.3,\Dy + 0.25 -0.5) {};
\node[amr] (nC4) at (\Dx-0.3, \Dy - 0.25 -0.5) {};
\node[amr] (nC5) at (\Dx+0.6,\Dy -0.5) {};
\node[amr] (nC6) at (\Dx+0.6,\Dy-0.25 -0.5) {};

\draw[->] (nA2) -- (nA1);
\draw[->] (nA2) -- (nB3);
\draw[->] (nB3) -- (nA4);
 
\draw[->] (nB1) -- (nB2);
\draw[->] (nB3) -- (nB2);
\draw[->] (nB3) -- (nB4);
\draw[->] (nB1) to [out=260,in=230] (nB4);

\draw[->] (nC2) to [out=130,in=210] (nB1);
\draw[->] (nC2) -- (nC4);
\draw[->] (nC3) -- (nC2);
\draw[->] (nC3) -- (nC5);
\draw[->] (nC5) -- (nC6);

\node[amr,fill=\matchcolorB] (nB1) at (\Ex-0.3, \Ey + 0.25) {};
\node[amr,opacity=\skipopacity] (nB2) at (\Ex, \Ey) {};
\node[amr,fill=\matchcolorA] (nB3) at (\Ex+0.3,\Ey + 0.25) {};
\node[amr,fill=\matchcolorA,opacity=\skipopacity] (nB4) at (\Ex+0.6, \Ey) {};
 
\node[amr,opacity=\skipopacity] (nA1) at (\Ex-0.3, \Ey+0.5) {};
\node[amr] (nA2) at (\Ex, \Ey + 0.75) {};
\node[amr,opacity=\skipopacity] (nA4) at (\Ex+0.6, \Ey + 0.75) {};
\node[amr] (nC2) at (\Ex, \Ey-0.5) {};
\node[amr,opacity=\skipopacity] (nC3) at (\Ex+0.3,\Ey + 0.25 -0.5) {};
\node[amr] (nC4) at (\Ex-0.3, \Ey - 0.25 -0.5) {};
\node[amr,opacity=\skipopacity] (nC5) at (\Ex+0.6,\Ey -0.5) {};
\node[amr,opacity=\skipopacity] (nC6) at (\Ex+0.6,\Ey-0.25 -0.5) {};

\draw[->,opacity=\skipopacity] (nA2) -- (nA1);
\draw[->] (nA2) -- (nB3);
\draw[->,opacity=\skipopacity] (nB3) -- (nA4);
 
\draw[->,opacity=\skipopacity] (nB1) -- (nB2);
\draw[->,opacity=\skipopacity] (nB3) -- (nB2);
\draw[->,opacity=\skipopacity] (nB3) -- (nB4);
\draw[->,opacity=\skipopacity] (nB1) to [out=260,in=230] (nB4);

\draw[->] (nC2) to [out=130,in=210] (nB1);
\draw[->] (nC2) -- (nC4);
\draw[->,opacity=\skipopacity] (nC3) -- (nC2);
\draw[->,opacity=\skipopacity] (nC3) -- (nC5);
\draw[->,opacity=\skipopacity] (nC5) -- (nC6);

%\draw (0,1) circle (1);
%\filldraw[fill=gray!50!white,even odd rule,densely dashed](0,1) circle (3) (0,2) circle (1);\draw [<->](0,-1.5) -- (0,3.5);\draw [<->](-2.5,0)--(2.5,0);\draw (-0.1,1)--(0.1,1);\node [scale=.8] at (0.2,1) {$i$};\node at (2.2, .2)[scale=0.7] {Re};\node at (-.3,3.3)[scale=0.7]{Im};
\end{tikzpicture}
\end{center}
\caption{The single document AMR summarization pipeline proposed by \citet{liu-etal-2015-toward-abstractive}. The colored nodes indicate concepts that are coreferent across sentences and merged into a single node in the document graph.% \nvnote{are the node colors supposed to mean anything?}
}
\label{fig:pipeline}
\end{figure}

Previous work on summarization with AMR has used different merging strategies to link nodes across sentence graphs in order to form a coherent document graph \citep{liu-etal-2015-toward-abstractive,hardy-vlachos-2018-guided,dohare-etal-2018-unsupervised}. However, the effects of the specific choices made about how to merge nodes have not been substantially explored. Thus, in this paper, we collect a dataset of annotations consisting of co-reference alignments between document and summary nodes which can be used to evaluate the specific performance of a merge strategy at each step of the content selection process. We also present a novel, person-focused merge strategy through the use of co-reference across sub-graphs, and demonstrate improvement upon the seminal approach to content selection using AMR introduced by \citet{liu-etal-2015-toward-abstractive}.

In summary, our contributions are as follows\footnote{We provide our annotation guidelines and a sample of our dataset in the supplementary data. We will publicly release the full version of the data and code for the merge strategies and evaluations upon publication.}:
\begin{itemize}
    \item Adjudicated annotations for node alignment between documents and summaries that can be used to evaluate the efficacy of document sentence merging and node-level content selection (\autoref{sec:annotations});
    \item A merge strategy combining aspects of \citet{liu-etal-2015-toward-abstractive}'s work with person-focused co-reference across sub-graphs (\autoref{sec:merging});
    \item An evaluation of our proposed merge strategy that %\nvnote{substantially?}
    demonstrates significantly improved performance on both node merging and content selection (\autoref{sec:experiments}).
\end{itemize}
\section{Related Work}
To the best of  our knowledge, \citet{ogorman-etal-2018-amr} is the only prior work providing a dataset for cross-sentence AMR node co-reference. The primary distinction between their annotations and ours is that while they annotate co-reference chains across sentences in individual documents, we focus specifically on AMR as applied to summarization, and explicitly annotate alignments between documents and summaries rather than within documents. This allows us to evaluate not only performance on within-document merging but on the additional step of node-level content selection as well. We apply these annotations to evaluate merging and content selection in existing methods of single-document AMR summarization.

Seminal work on summarization using AMR~\cite{liu-etal-2015-toward-abstractive} first merges the AMR annotations for all sentences in a particular document into a single graph by identifying nodes with identical text labels, then treats summarization of the document as a subgraph selection task and solves it using integer linear programming. They evaluate their selected subgraph by measuring overlap of nodes selected from the merged graph with labels inferred from alignments to the summary nodes. Subsequent research in this space builds on or compares to this approach. For example, \citet{dohare-etal-2018-unsupervised} suggest an alternative way of merging the document AMR into a single graph using text coreference, but their method merges nodes associated with the key word in co-referent noun phrases, while we merge nodes by focusing on components of the graph. Additional work on AMR summarization follows that of \citet{liu-etal-2015-toward-abstractive} for merging but instead focuses on \textit{multi-document} summarization~\cite{liao-etal-2018-abstract} or methods for natural language  realization of the summary AMR graph~\cite{hardy-vlachos-2018-guided}. In contrast, our focus is on content selection for single-document summarization.
\section{Data}
\subsection{AMR Corpus}
We use release 3.0 of the AMR annotation corpus \cite{amrbank}, which consists of sections from a variety of domains including blogs, Aesop's Fables, Wikipedia, and newswire text. We specifically focus on the proxy report section, which is the only section in the corpus with AMR for document-summary pairs rather than standalone documents. Each document in this section is a news article paired with a human-written abstractive summary, or ``proxy report". As in the other sections of the corpus, both documents and summaries are annotated with AMR graphs for each sentence.

The proxy section contains 298 document-summary pairs in the training set, one of which we discard because the associated summary is empty; 35 in the development set; and 33 in the test set. The average document length is 16.9 sentences, with a standard deviation of 9.1; the average summary length is 1.6 sentences.

The AMR graph for each sentence is a directed graph with labels on nodes and edges. Every node in the graph is associated with a concept that may be either a noun or a verb sense, and is assigned a text label representing that concept. Edges are directed according to the relation they represent, and are labeled with one of a fixed set of relation types between nodes.

\subsection{Annotations}
\label{sec:annotations}
Our annotations are pairings between document and summary nodes labeling co-reference between those nodes.

We annotated 50 document-summary pairs in total, including the full 33 documents from the test set as well as an additional 17 from the development set. We recruited ten students in total as annotators, all undergraduate or graduate students with a background in NLP, nine of whom are native English speakers. Each document was assigned to two annotators who first performed the annotation task independently, then worked together to adjudicate disagreements. A sample of the final adjudicated labels and the IDs of the associated documents in the original corpus are provided as supplementary data. We include further annotation details in \autoref{app:interface}, and full guidelines for the annotation task in the supplementary material.

\paragraph{Agreement and adjudication}
As traditional agreement metrics such as Cohen's $\kappa$ are not applicable to our setting, which may be viewed as a binary multi-label classification problem, we instead compute two set-based metrics for annotator agreement over the set of nodes which received at least one alignment from either annotator. The first is based on exact match: the label sets given by two annotators are considered a match if they are precisely identical. The second is the softer notion of Jaccard similarity between the two label sets.

The preliminary round of annotations achieved .487 exact match agreement and .511 average Jaccard similarity, both of which are far higher than the expected values from random annotation, given that each document node may be aligned to on the order of 20-30 summary entities. However, as our annotations are designed to be a relatively small but high-quality evaluation set, we performed an adjudication round in which the pair of annotators assigned to each document worked together to resolve any disagreements, yielding a finalized set of adjudicated gold labels for each node.
\newcommand{\concepts}{\mathcal{V}}
\newcommand{\relations}{\mathcal{E}}
\newcommand{\snodes}{V}
\newcommand{\sedges}{E}
\newcommand{\innodes}{V_:}
\newcommand{\inedges}{E_:}
\newcommand{\doc}{X}
\newcommand{\dsize}{n}
\newcommand{\mergestrat}{M}
\newcommand{\inspace}{\mathcal{X}}
\newcommand{\outspace}{???}
\newcommand{\dnodes}{\Lambda}
\newcommand{\dedges}{\xi}
\newcommand{\mnode}{\nu}
\newcommand{\cluster}{C}
\newcommand{\sumnodes}{S}
\newcommand{\sumedges}{R}
\newcommand{\nlabel}{y}
\newcommand{\prednlabel}{\hat{\nlabel}}
\newcommand{\clf}{f}
\newcommand{\params}{\theta}

\section{Methods}
\label{sect:methods}

Following \citet{liu-etal-2015-toward-abstractive}, we model AMR summarization as a three
stage pipeline (depicted in \autoref{fig:pipeline}) of \textit{(i)} sentence graph combination, \textit{(ii)} content selection, and \textit{(iii)} summary text generation.

\paragraph{Sentence Graph Combination} The input to the pipeline is an ordered sequence of $\dsize$ disjoint 
AMR graphs,  $\left(\snodes_1, \sedges_1\right),
\ldots,
\left(\snodes_\dsize,\sedges_\dsize  \right)$, where $\snodes_i$ and $\sedges_i$
are the set of nodes and edges respectively of the AMR graph representation of sentence $i$.\footnote{Technically, the $\snodes_i$ are multi-sets since they 
can contain two or more  distinct instances of the same concept (this happens in \autoref{fig:merge}, where there are multiple instances of \textit{person}, \textit{country}, and other concepts).} A node $v \in \snodes_i$ corresponds to an instance of an AMR concept and an edge $(v,v^\prime)$ exists in $\sedges_i$ if $v^\prime$ occupies a role in the sentence with respect to $v$ (e.g., if $v^\prime$ is the ARG0 of $v$). In this stage of the pipeline, the disjoint sentence graphs are connected into a connected graph $\left(\dnodes, \dedges \right)$ representing the semantics of the entire document. We describe several ways of automatically performing sentence graph merging in \autoref{sec:merging}. 

\paragraph{Node Selection} We develop a model to identify a summary subgraph $\left(\sumnodes, \sumedges\right) \subseteq \left(\dnodes, \dedges\right)$ corresponding to the AMR graph of the summary text. We treat this as a node-level binary classification task, predicting for each node $v\in\dnodes$ whether to include it in $\sumnodes$ or not. Noisy training labels for this task are inferred from the summary graphs by assigning a label of 1 to a document node if there is at least one summary node with the same concept label, and 0 otherwise. We train a graph attention network (GAT) \cite{velickovic2018graph} to perform the classification. The edges of the summary subgraph are determined implicitly by the summary nodes (i.e., $\sumedges = \left\{(v,v^\prime) \in \dedges | v,v^\prime \in \sumnodes   \right\}).$ Additional details about the GAT model can be found in \autoref{app:content}.

\paragraph{Summary Text Generation} 
Finally, a natural language generation (NLG) model is tasked with 
mapping $\left(\sumnodes, \sumedges\right)$ to a natural language summary. We fine-tune BART \citep{lewis-etal-2020-bart} on the training split of the proxy corpus to generate summary text from a linearization of the selected nodes.

\subsection{Sentence Graph Combination}
\label{sec:merging}

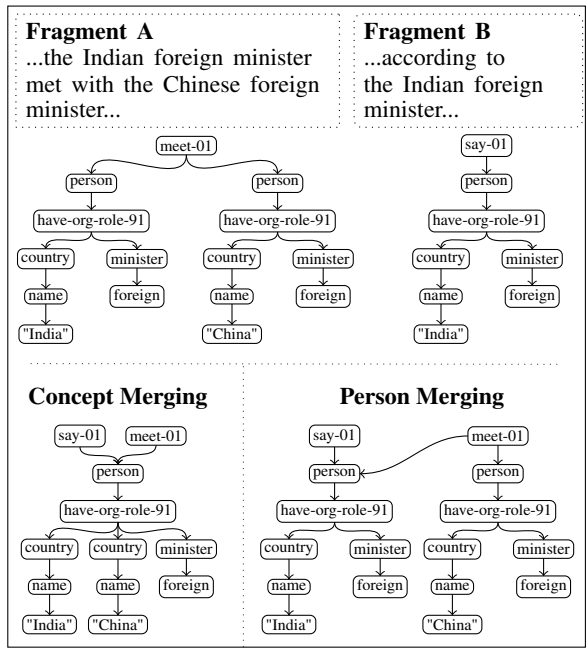
\begin{figure}[t]
\begin{center}
\usetikzlibrary{shapes}
\begin{tikzpicture}[scale=1.0,
amr/.style={draw,inner sep=0.5mm, outer sep=0pt, rounded corners=0.75mm},
arc/.style={->},
]
\draw (-3.8,4.5) rectangle (3.8,-4);
\def\sAx{-2.70}
\def\sAy{2.15}
\def\sBx{-0.25}
\def\sCx{2.5}
\def\sDx{-2.35}
\def\sDy{-1.7}
\def\perAx{0.50}
\def\perBx{2.65}
\node[amr] (nA1) at (\sAx+1.25,\sAy+0.5) {\tiny meet-01};
\node[amr] (nA2) at (\sAx,\sAy) {\tiny person};

\node[amr] (nA3) at (\sAx,\sAy-0.5) {\tiny have-org-role-91};
\node[amr] (nA4) at (\sAx+0.6,\sAy-1.0) {\tiny minister};
\node[amr] (nA5) at (\sAx+0.6,\sAy-1.5) {\tiny foreign};
\node[amr] (nA6) at (\sAx-0.6,\sAy-1.0) {\tiny country};
\node[amr] (nA7) at (\sAx-0.6,\sAy-1.5) {\tiny name};
\node[amr] (nA8) at (\sAx-0.6,\sAy-2.0) {\tiny "India"};
\draw[arc] (nA1) to [out=270,in=90] (nA2);
\draw[arc] (nA2) to (nA3);
\draw[arc] (nA3) to [out=270,in=90] (nA4);
\draw[arc] (nA4) to (nA5);
\draw[arc] (nA3) to [out=270,in=90] (nA6);
\draw[arc] (nA6) to (nA7);
\draw[arc] (nA7) to (nA8);

\node[amr] (nB1) at (\sBx,\sAy) {\tiny person};
\node[amr] (nB2) at (\sBx,\sAy-0.5) {\tiny have-org-role-91};
\node[amr] (nB3) at (\sBx+0.6,\sAy-1.0) {\tiny minister};
\node[amr] (nB4) at (\sBx+0.6,\sAy-1.5) {\tiny foreign};
\node[amr] (nB5) at (\sBx-0.6,\sAy-1.0) {\tiny country};
\node[amr] (nB6) at (\sBx-0.6,\sAy-1.5) {\tiny name};
\node[amr] (nB7) at (\sBx-0.6,\sAy-2.0) {\tiny "China"};

\draw[arc] (nA1) to [out=270,in=90] (nB1);
\draw[arc] (nB1) to (nB2);
\draw[arc] (nB2) to [out=270,in=90] (nB3);
\draw[arc] (nB3) to (nB4);
\draw[arc] (nB2) to [out=270,in=90] (nB5);
\draw[arc] (nB5) to (nB6);
\draw[arc] (nB6) to (nB7);

\node[amr] (nC1) at (\sCx,\sAy+0.5) {\tiny say-01};
\node[amr] (nC2) at (\sCx,\sAy) {\tiny person};
\node[amr] (nC3) at (\sCx,\sAy-0.5) {\tiny have-org-role-91};
\node[amr] (nC4) at (\sCx+0.6,\sAy-1.0) {\tiny minister};
\node[amr] (nC5) at (\sCx+0.6,\sAy-1.5) {\tiny foreign};
\node[amr] (nC6) at (\sCx-0.6,\sAy-1.0) {\tiny country};
\node[amr] (nC7) at (\sCx-0.6,\sAy-1.5) {\tiny name};
\node[amr] (nC8) at (\sCx-0.6,\sAy-2.0) {\tiny "India"};
\draw[arc] (nC1) to (nC2);
\draw[arc] (nC2) to (nC3);
\draw[arc] (nC3) to [out=270,in=90] (nC4);
\draw[arc] (nC4) to (nC5);

\draw[arc] (nC3) to [out=270,in=90] (nC6);
\draw[arc] (nC6) to (nC7);
\draw[arc] (nC7) to (nC8);

\node[amr] (nT1) at (\sDx-0.5,\sDy+0.5) {\tiny say-01};
\node[amr] (nT2) at (\sDx+0.5,\sDy+0.5) {\tiny meet-01};
\node[amr] (nT3) at (\sDx,\sDy) {\tiny person};
\node[amr] (nT4) at (\sDx,\sDy-0.5) {\tiny have-org-role-91};
\node[amr] (nT5) at (\sDx+0.9,\sDy-1.0) {\tiny minister};
\node[amr] (nT6) at (\sDx+0.9,\sDy-1.5) {\tiny foreign};
\node[amr] (nT7) at (\sDx-0.9,\sDy-1.0) {\tiny country};
\node[amr] (nT8) at (\sDx-0.9,\sDy-1.5) {\tiny name};
\node[amr] (nT9) at (\sDx-0.9,\sDy-2.0) {\tiny "India"};
\node[amr] (nT10) at (\sDx,\sDy-1.0) {\tiny country};
\node[amr] (nT11) at (\sDx,\sDy-1.5) {\tiny name};
\node[amr] (nT12) at (\sDx,\sDy-2.0) {\tiny "China"};

\draw[arc] (nT1) to [out=270,in=90] (nT3);
\draw[arc] (nT2) to [out=270,in=90] (nT3);
\draw[arc] (nT3) to (nT4);
\draw[arc] (nT4) to [out=270,in=90] (nT5);
\draw[arc] (nT5) to (nT6);

\draw[arc] (nT4) to [out=270,in=90] (nT7);
\draw[arc] (nT7) to (nT8);
\draw[arc] (nT8) to (nT9);

\draw[arc] (nT4) to [out=270,in=90] (nT10);
\draw[arc] (nT10) to (nT11);
\draw[arc] (nT11) to (nT12);

\node[amr] (perA1) at (\perAx,\sDy+0.5) {\tiny say-01};
\node[amr] (perA2) at (\perBx,\sDy+0.5) {\tiny meet-01};
\node[amr] (perA3) at (\perAx,\sDy) {\tiny person};
\node[amr] (perA4) at (\perAx,\sDy-0.5) {\tiny have-org-role-91};
\node[amr] (perA5) at (\perAx+0.6,\sDy-1.0) {\tiny minister};
\node[amr] (perA6) at (\perAx+0.6,\sDy-1.5) {\tiny foreign};
\node[amr] (perA7) at (\perAx-0.6,\sDy-1.0) {\tiny country};
\node[amr] (perA8) at (\perAx-0.6,\sDy-1.5) {\tiny name};
\node[amr] (perA9) at (\perAx-0.6,\sDy-2.0) {\tiny "India"};

\node[amr] (perB1) at (\perBx,\sDy) {\tiny person};
\node[amr] (perB2) at (\perBx,\sDy-0.5) {\tiny have-org-role-91};
\node[amr] (perB3) at (\perBx+0.6,\sDy-1.0) {\tiny minister};
\node[amr] (perB4) at (\perBx+0.6,\sDy-1.5) {\tiny foreign};
\node[amr] (perB5) at (\perBx-0.6,\sDy-1.0) {\tiny country};
\node[amr] (perB6) at (\perBx-0.6,\sDy-1.5) {\tiny name};
\node[amr] (perB7) at (\perBx-0.6,\sDy-2.0) {\tiny "China"};
\draw[arc] (perA1) to (perA3);
\draw[arc] (perA3) to (perA4);
\draw[arc] (perA4) to [out=270,in=90] (perA5);
\draw[arc] (perA5) to  (perA6);
\draw[arc] (perA4) to [out=270,in=90] (perA7);
\draw[arc] (perA7) to  (perA8);
\draw[arc] (perA8) to  (perA9);

\draw[arc] (perA2) to (perB1);
\draw[arc] (perA2) to [out=180,in=0] (perA3);

\draw[arc] (perB1) to  (perB2);
\draw[arc] (perB3) to (perB4);
\draw[arc] (perB2) to [out=270,in=90] (perB3);
\draw[arc] (perB2) to [out=270,in=90] (perB5);
\draw[arc] (perB5) to  (perB6);
\draw[arc] (perB6) to  (perB7);

\node at (\sDx,\sDy+1.0) {\small \textbf{Concept Merging}};
\node at (\perAx+1.15,\sDy+1.0) {\small \textbf{Person Merging}};

\draw[-,dotted,opacity=0.75] (-0.7,-4) -- (-0.7,-0.25);
\draw[-,dotted,opacity=0.75] (-3.5,-0.24) -- (3.5,-0.25);

% \node[align=left,text width=4cm,text height=0.1mm] at (\sAx+1.25, \sAy+2.25) {\small \textbf{Sentence Fragment A}};
% \node[align=left,text width=4cm,text height=0.1mm] at (\sAx+5.25, \sAy+2.25) {\small \textbf{Sentence Fragment B}};

\node[anchor=north west,align=left,text width=4cm,text height=2mm,draw,dotted,rounded corners=0.1mm] at (\sAx-1, \sAy+2.25) {\small \textbf{Fragment A}\\ \small ...the Indian foreign minister\\\small  met with the Chinese foreign\\[-4.0pt]
\small minister...};

\node[anchor=north west,align=left,text width=2.75cm,text height=2mm,draw,dotted,rounded corners=1mm] at (\sCx-1.75, \sAy+2.25) {\small \textbf{Fragment B}\\\small ...according to\\\small the Indian foreign\\[-4pt] 
minister...};

%\node[amr] at (\sAx+0.7,\sAy-0.5) {\tiny person};

\end{tikzpicture}
\end{center}
\caption{Example of Concept Merging versus Person Merging. Note that in this instance, Concept Merging collapses two distinct individuals into a single \textit{person} node while Person Merging is able to preserve this information.}
\label{fig:merge}
\end{figure}

We assume each sentence of each input document is annotated with its own AMR graph.\footnote{In this work we use the gold AMRs graphs provided by the AMR corpus but they could in principal be provided by an AMR parser.} This means that while multiple references to the same instance of a concept are annotated as such within sentences (e.g., if the same person is mentioned multiple times in a sentence,
they will only have a single \textit{person} node in the corresponding AMR graph), 
they are not annotated for co-reference across sentences.

Performing content selection directly 
on this disjoint collection of AMR graphs is not
ideal. First, under a strict interpretation of AMR, distinct nodes correspond to distinct instances, which forces the content selection model to consider each mention
of a concept in isolation, making it difficult to
capture how the same instance of a concept might
participate in many other events in a document. 

We hypothesize that an ideal document AMR 
graph would contain a single node 
for each distinct instance of a concept mentioned
in the sentence AMR graphs. In other words, 
nodes that were co-referent across sentence AMRs
would be ``merged'' into a single node in the document AMR. This merged node would inherit the incoming and outgoing edges to 
other concepts in the original AMRs, and  thus
frequently occuring instances in the document 
would take on greater graph centrality in the document AMR.
Centrality is a useful feature for summarization using lexically constructed graphs \cite{erkan04lexrank}, and we suspect this to hold for semantically constructed graphs as well.

We explore three methods of merging nodes to produce a document AMR graph, which we refer to as Concept Merging, Person Merging, and Combined Merging.\footnote{We also considered evaluating the merge strategy outlined in \citet{dohare-etal-2018-unsupervised}, but were unable to retrieve the node merges from their published code, and so we do not compare against their method in this paper.}

\paragraph{Concept Merging}
\citet{liu-etal-2015-toward-abstractive} merge any sentence-level nodes into a single node in
the document graph if they share the same AMR concept regardless of whether they are different instances. 
Formally, 
\citeauthor{liu-etal-2015-toward-abstractive} partition the sentence nodes into disjoint sets $\cluster_j$ such that $ \bigcup_{i=1}^\dsize \snodes_i = \bigcup_{j=1}^m \cluster_j$ and
$\operatorname{concept}(c) = \operatorname{concept}(c^\prime)$ for
any $c,c^\prime \in \cluster_j$.
For each $\cluster_j$, a single node $\mnode$ of the same concept is created in  
$\dnodes$. % while the nodes in $\cluster$ are not added to $\dnodes.$ 
For any $\mnode_i,\mnode_j \in \dnodes$ with corresponding concept clusters $\cluster_i$ and $\cluster_j$ respectively, there exists an edge $(\mnode_i, \mnode_j) \in \dedges$
in the document graph if there exists a pair of 
sentence nodes $c_i \in \cluster_i$ and  $c_j \in \cluster_j$ with an edge $(c_i, c_j) \in \bigcup_{k=1}^n \sedges_k$.

The primary flaw of this strategy is the fact that, with some 
exceptions (due to an initial step collapsing name and date nodes,
which we elaborate upon in \autoref{sec:error-analysis}),
the resulting document AMR has no way of distinguishing between multiple instances of the 
same type of concept. For example, in many cases, nodes representing different instances of 
the \textit{person} concept will be merged across sentences without regard to whether those nodes actually refer to the same person or not. As an example, see the Concept Merging example in \autoref{fig:merge} where two different people (the Indian and Chinese foreign ministers)
are collapsed to a single person node.

\paragraph{Person Merging}
Our first strategy for merging is inspired by the observation that many of the merge errors in concept merging involve incorrect merging of nodes that occur in the descendant subgraph of a specific instance of some entity -- for example, identifying nodes that are attached to a particular person. To address this issue, we take a person-focused approach to merging as follows: 
\begin{enumerate}
\item Identify the subgraph consisting of all descendant nodes of every node representing the \textit{person} concept.
\item Identify the text spans that correspond to each of these nodes using automatic AMR-to-text alignment.
\item Use a textual co-reference system to determine whether any of the text spans corresponding to each of any two sub-graphs are co-referent; and if so, determine those two sub-graphs to be co-referent.
\item For any two co-referent person sub-graphs, merge any nodes across the two sub-graphs with the same node label.
\end{enumerate}
   We will refer to these graphs as ``person-merged.'' An example of person merging can be found in \autoref{fig:merge}, where both distinct instances of the \textit{person} concept along with their sub-trees are preserved.  

\paragraph{Combined Merging}
Our final merge strategy combines both Concept and Person merging. In this joint strategy, we first perform Person Merging. We then perform concept-merging on all remaining nodes of the sentence graphs (but do not perform an initial name or date collapse), excluding any person nodes or their subtrees that have already been merged in the Person Merging phase.
Detailed pseudocode for the combined merging method can be found in \autoref{app:person}.
\section{Experiments}
\label{sec:experiments}
In our experiments we evaluate the merge methods laid out in the previous section. We additionally consider an unmerged baseline where sentences are simply joined at the root by an artificial root node.

We evaluate these merge styles in two ways: we compare the proposed merge clusters directly against those induced by our human annotations using coreference metrics (node merging); and we evaluate the summary nodes predicted by a classifier trained on each style of merged graph against our annotated gold alignments with precision, recall and F-measure over the unmerged sentence nodes aligned to reference summary  nodes (node selection).

\subsection{Data handling and merging}
We use the TAMR aligner \cite{liu-etal-2018-amr} to generate node-to-text alignments. We use the Penman \cite{goodman-2020-penman} and NetworkX packages in Python to read and store the resulting AMR graph structures, respectively. For text coreference we use SpanBERT-large \cite{joshi-etal-2020-spanbert}.
Regardless of merge method, we also join sentences at the root with an artificial root node when forming the document graph.

\subsection{Evaluating merge clusters}
Each merge strategy produces a partition of the
sentence-level AMR graph. Given this partition, we may consider the associated co-reference cluster for each node of the graph to be the set of other nodes which occur within the same element of the partition, and can evaluate these clusters against those we derive from the annotations.

Although our collected annotations are for document-summary alignment rather than document merging, it is simple to infer merge clusters from these alignments: every document node can be said to be co-referent with all other document nodes that are aligned to one of the same summary nodes. In this manner we can collect a set of gold merge clusters that we may evaluate our automatic clusters against. We note that this approach does mean that any nodes not present in the summary will be treated as unmerged, but for summarization purposes we are only interested in 
summary-aligned nodes.

\begin{table*}[t]
    \centering
        %\resizebox{0.49\textwidth}{!}{
    \begin{tabular}{cc c cc c c}
        \toprule
         & \multicolumn{3}{c}{$B^3$} & \multicolumn{3}{c}{LEA}\\
         \cmidrule(lr){2-4}\cmidrule(lr){5-7}
         
         Method & Prec. & Recall & F1 & Prec. & Recall & F1 \\ \midrule
        Person & 0.011 & 0.023 & 0.015 & 0.052 & 0.144 & 0.074 \\
        Concept*$\dagger$ & 0.469 & 0.514 & 0.490 & \textbf{0.088} & 0.191 & 0.115 \\
        Combined*$\dagger$ & \textbf{0.743} & \textbf{0.704} & \textbf{0.723} & 0.086 & \textbf{0.241} & \textbf{0.122} \\ \bottomrule
    \end{tabular}%}
    \caption{Precision, recall, and F1 for the $B^3$ and LEA metrics for person merging, concept merging~\citep{liu-etal-2015-toward-abstractive}, and our full approach (combined merge). Methods are marked with  (*) if the $B^3$ F-score is statistically significant over that of the row directly above, and with ($\dagger$) if the LEA F-score is significant over the row above.}
    \label{tab:merge_scores}
\end{table*}

Although the co-reference task we are interested in is over nodes rather than text, it shares with text co-reference the objective of determining which semantic objects appear in the same cluster. We thus evaluate these clusters using two textual co-reference metrics, $B^3$ \cite{Bagga98algorithmsfor} and LEA \cite{moosavi2016lea}, implemented over our AMR data structure. We do not use importance weighting for the latter.

\subsection{Node selection}
In the node selection phase, the merged or unmerged document graph produced in the previous step is given as input to a node classifier which learns to predict a binary label for each node indicating whether it should be present in the summary. Here we follow prior work by using label-based alignment between documents and summaries to generate the training labels - that is, a document node is assigned a ``gold" label of 1 for training purposes if there is at least one node in the summary with the same concept label, and 0 otherwise. This is an imperfect method for the same reasons that apply to the merging stage; while these noisy labels may be sufficient for training, we therefore need an external benchmark to objectively evaluate content selection with respect to the true summary.

Thus, we use our annotations as gold standard here as well: we consider the true label of a node to be 1 if it is aligned to at least one node in the summary in our annotations, and 0 otherwise. To avoid introducing confounding effects by counting merged nodes once in a merged graph but multiple times in the unmerged graph, we propagate the labels predicted on each node in a merged document graph back to the original sentence nodes that formed it, and compute precision, recall and F-measure against the gold labels induced by the alignments over the unmerged nodes.

We use the same GAT architecture and training procedure to perform node classification on all merge strategies. The details of both are described in \autoref{app:content}.

\subsection{Generation}
Finally, to generate summary text from the nodes selected with each merge strategy, we linearize the extracted nodes of the graph into a sequential format. This linearized AMR is then passed to a pretrained BART-large model which has been finetuned to generate the summary text from such linearized input. We include further details on the linearization and finetuning procedure in \autoref{app:finetuning}.

We compare the generated output from each merge strategy using three kinds of automatic metrics (ROUGE \citep{lin2004rouge}, METEOR \citep{banerjee-lavie-2005-meteor}, and MoverScore \citep{zhao-etal-2019-moverscore}), as well as with a human evaluation designed to investigate the influence of our proposed merge strategies on person-related output, ranking each set of summaries on three criteria (fluency, salience and faithfulness with regards to information about humans). We provide sample summary outputs in \autoref{app:finetuning}, and report details of the human evaluation setup in \autoref{app:humaneval}.
\section{Results}
In this section we report results for node merging and node selection. For node selection, we report the average of precision, recall and F-score over five classifier training runs in order to control for random variation in the model initialization. For statistical significance  we use the Approximate Randomization Test \cite{riezler05}.

\subsection{Node merging}
We report precision, recall and F-score for the $B^3$ and LEA metrics over the node clusters induced by person merging, concept merging and combined merging, as compared to the gold annotated clusters (see \autoref{tab:merge_scores}).

Interestingly, combined merging strongly outperforms both person and concept merging on nearly all metrics. While it is in line with expectations that cluster precision would be higher for our method than for concept merging, as it is more selective about which nodes may be merged, it is surprising that our method also yields substantially higher recall. We discuss this phenomenon in \autoref{sec:analysis-person}.

\subsection{Node selection}
\begin{table}[h]
    \centering
    %\resizebox{0.49\textwidth}{!}{
    \begin{tabular}{c c c c}
    \toprule
    Merge Strategy  &  Prec. & Recall & F1\\
    \midrule
    Person       & 0.348 & 0.244 & 0.280\\
    Concept (\citeauthor{liu-etal-2015-toward-abstractive})*        & 0.363 & 0.277 & 0.293\\
    Unmerged*    & 0.327 & 0.359 & 0.332\\
    Combined*      & \textbf{0.412} & \textbf{0.400} & \textbf{0.398}\\
    %Person+Unmerged & 0.289 & 0.469 & 0.350 \\
    \bottomrule
    \end{tabular}%}
    \caption{Precision, recall, and F1 of document node selection under different merge strategies. A method is marked with (*) if the difference between it and the method one row above is statistically significant.}
    \label{tab:content_selection}
\end{table}

% unmerged	0.3833	0.3209	0.3419
% type	0.3656	0.1491	0.2068
% person	0.587	0.2516	0.3431
% coref only	0.3526	0.2576	0.2888

% multiply prediction values for plain + mix	0.3277	0.5442	0.3978

We evaluate the unmerged and concept merging baselines, as well as both our combined merging strategy and person merging alone. Results are presented in  \autoref{tab:content_selection}. Combined merging's performance in node clustering seems to effectively carry over to node selection, where it continues to outperform concept merging and person merging in both precision and recall.

\subsection{Generation}
\begin{table*}[]
    \centering
    \begin{tabular}{c|c c c c c}
        \toprule
        Merge & ROUGE-1 & ROUGE-2 & ROUGE-L & METEOR & MoverScore \\
        \midrule
        Unmerged & 0.339 & 0.098 & 0.230 & 0.250 & 0.240 \\
        Concept & 0.284 & 0.066 & 0.210 & 0.175 & 0.211 \\
        Combined & 0.305 & 0.071 & 0.210 & 0.180 & 0.224 \\ \midrule
        Seq-to-seq BART (CNN) & 0.417 & 0.237 & 0.330 & 0.288 & 0.390 \\
        \bottomrule
    \end{tabular}
    \caption{Scores on automatic metrics for each merge strategy, and the pretrained sequence-to-sequence BART-large model finetuned on CNN/DailyMail. Reported ROUGE numbers are f-scores.}
    \label{tab:generation}
\end{table*}

\begin{table}[]
    \centering
    \begin{tabular}{c|c c c}
        \toprule
        Merge & Fluency & Salience & Faithfulness \\
        \midrule
        Unmerged & 20 (8) & \textbf{21 (8)} & 16 (6) \\
        Concept & 21 (10) & 15 (6) & 15 (5) \\
        Combined & \textbf{22 (8)} & 18 (3) & \textbf{17 (7)} \\
        \bottomrule
    \end{tabular}
    \caption{Number of test set documents for which each merge strategy was ranked first by at least one annotator (both annotators) on each criterion. The test set contains 33 documents, but the numbers do not sum to 33 because ties are permitted.}
    \label{tab:generation-human}
\end{table}

We report the performance of our basic finetuned BART models when given as input the linearized selected AMR from the unmerged, concept, and combined merge strategies on
a range of automatic metrics in \autoref{tab:generation}.\footnote{While we also provide the scores for the pretrained sequence-to-sequence BART model finetuned on CNN/DailyMail as a performance ceiling, we note that our AMR-to-text BART models are not intended to provide summary output competitive with end-to-end BART, but are rather a diagnostic tool to investigate the effects that the different merge strategies have upon the final generated output. Our AMR-to-text models were finetuned on an extremely small amount of data and should not be considered fully pretrained for this task; a significant amount of further work on tuning would need to be done in order to make a comparison between AMR-to-text and sequence-to-sequence BART models.} We report the performance of our finetuned BART models with the same three merge strategies under human evaluation for fluency, salience and faithfulness in \autoref{tab:generation-human}.

While BART outputs using the unmerged strategy appear superior under all automatic metrics, the human evaluation reveals a different array of strengths, in which unmerged outputs are rated the best in terms of salience the most often by far, but the combined merge strategy actually yields the most faithful outputs in general.
\section{Analysis}
\begin{table*}[t]
    \centering
    \begin{tabular}{c|p{10cm}}
        \toprule Error type & Example sentence fragment pair that produces error\\ \midrule
        \multirow{2}{*}{Concept conflation (surface)} & ...required to inform any person whose \textbf{assets} are being frozen... \\
        & ...an European Union decision to freeze the \textbf{assets} of the People's Mujahadeen of Iran.\\ \hline
        \multirow{2}{*}{Concept conflation (hidden)} & ...the foreign \textbf{minister} of India... \\
        & ...the foreign \textbf{minister} of China...\\\bottomrule
    \end{tabular}
    \caption{Examples of concept conflation errors in concept merging.}
    \label{tab:small-errors}
\end{table*}

\subsection{Concept error analysis}
We identify nodes that concept merging clustered incorrectly using our annotated alignments and perform a manual inspection of concept merging errors on the development set, identifying five types of common errors: stopword conflation, missing synonyms, name skipping, conflation of surface concepts and conflation of hidden concepts. We present examples of the latter two kinds of error in \autoref{tab:small-errors}, and examples of all types of error in \autoref{sec:error-analysis}.

While the first two types of errors can be resolved with straightforward fixes (e.g., holding out stopwords when merging), and the third and fourth can be resolved using text-based coreference, the final and most complex type of error involves incorrectly merging AMR nodes that do not explicitly appear in the text, and thus cannot be resolved simply by performing coreference on the text itself. This error often occurs when there are multiple entities in the text of a concept type that induces a large implicit structure, such as the foreign ministers in \autoref{fig:merge}. Hidden concept conflation requires linking text coreference to the graph structure of the AMR, as the person-focused and combined merge strategies are designed to do.

\subsection{The role of coreference}
\label{sec:analysis-person}
Perhaps the most curious observation from these results is that while person merging has worse performance in both the cluster evaluation and on node selection than either unmerged or concept merged graphs - in fact, it has strikingly poor scores under merge cluster metrics - using it together with concept merging in the combined strategy yields by far the best results. This suggests that these two methods have strengths that are complementary.

\begin{table}[]
    \centering
    \begin{tabular}{c c}
        \toprule
        Method & Proportion of merged nodes\\
        \midrule
        Unmerged & 0\\
        Concept & 0.509 \\
        Person & 0.012 \\
        Combined & 0.572 \\
        \bottomrule
    \end{tabular}
    \caption{Average proportion of document nodes that are merged for each strategy, over the development set.}
    \label{tab:merge_proportion}
\end{table}

When we examine the clusters generated by person merging, we find that much of the poor performance can simply be attributed to the fact that it hardly merges any nodes at all (see \autoref{tab:merge_proportion}). This is unsurprising: there are relatively few nodes that occur both under a person node and within a coreference cluster, and even fewer that share a label with another person-descendant node in the same coreference cluster.

In this case, the benefit of using person merging as a first step in conjunction with
concept merging is not so much that it itself brings higher-quality merges, but rather that it takes a number of low-quality merges out of consideration by removing them from the pool of as yet unmerged nodes.

We further note that since the decision to merge nodes can only be made between nodes of the same label in the combined strategy, with additional restrictions between nodes in person subtrees that do not co-refer, we would expect its recall to be no higher than that of concept merging, which merges nodes with the same label without co-reference-based restriction.

The fact that there is also a large improvement in recall indicates that concept merging misses a sizable proportion of merges because of some condition that it does not share with combined merging. The only such difference is the initial name and date collapse step. This means that adding an identifying name to a node's label indeed has the adverse effect of preventing it from being merged with other instances of the same entity that were not similarly referred to by name in many cases, as described in \autoref{sec:error-analysis}.

Thus, combined merging also averts this type of error by replacing the name collapse step with co-reference-based subtree matching.

\subsection{Generated summaries}
The automatic metric scores indicate that while the combined merge strategy is slightly better than concept merging, the unmerged strategy outperforms both of the others. This is fairly surprising, as there does not seem to be any intuitive reason that leaving the AMR unmerged should be better for generation than merging it correctly. However, a manual inspection of the validation set reveals that the unmerged strategy generally yields much longer linearized inputs to the generator, and in fact on the validation set simply passing in the linearized AMR for the entire document graph yields better scores on automatic metrics than using the selected nodes from any merge strategy; we hypothesize that BART is powerful enough to identify redundant information even in AMR, which seemingly obviates the merging step.

However, the human evaluation tells a more nuanced story. While unmerged inputs yield the most salient outputs - supporting the hypothesis that BART performs its own internal pruning of redundant information - using the combined merge strategy improves faithfulness on person-related information, which suggests that narrowing down the field of consideration in the input helps it focus better on the actual information it is supposed to summarize rather than adding in details to fill in the gaps. The merge strategies are roughly evenly matched on fluency, but combined merge also has a slight edge here.
\section{Conclusion}
In this paper, we have presented a new annotated dataset of node alignments between document-summary AMR pairs that can be used to evaluate two fundamental components of the AMR-based summarization pipeline: document graph construction via node merging, and node-level content selection over the constructed document graph. Drawing on insights from error analysis of prior work when evaluated against our annotations, we developed a new method for node merging that combines co-reference with concept merging. We show that our novel method significantly outperforms prior work using concept merging alone when evaluated against our gold labels in both the node merging and node selection steps, and analyze the causes of error that our method addresses.

The most obvious direction for future work would be to focus on the generation component.
Another extension would be to use the interpretability our dataset affords to investigate the use of intermediate AMR representations for controllability in neural summarization. All in all, our work lays the foundation for a fine-grained understanding of content selection dynamics in AMR summarization.

\bibliographystyle{acl_natbib}
\bibliography{references}

\clearpage
\newpage
\appendix

\section{Annotation interface}
\label{app:interface}
Annotators are given the full list of summary sentences, followed by up to 10 document sentences for each document-summary pair.\footnote{Since this is single-document news summarization, the information in the summary typically occurs very early in the document.} For each sentence, they are presented with the sentence text first, followed by the list of entities that occur in that sentence, each represented by a handle and a label. The handle for each entity in a document sentence consists of the index of the node within the document (i.e., the handle of the third node of a sentence is `3' for that sentence). The handle for each entity in a summary sentence consists of the index of the sentence within the summary, and a unique identifier within the sentence (for example, the handle of the fourth node in the second summary sentence would be `2d'). The label provided for the entity is its concept label.

\begin{figure*}
    \centering
    \includegraphics[width=6in]{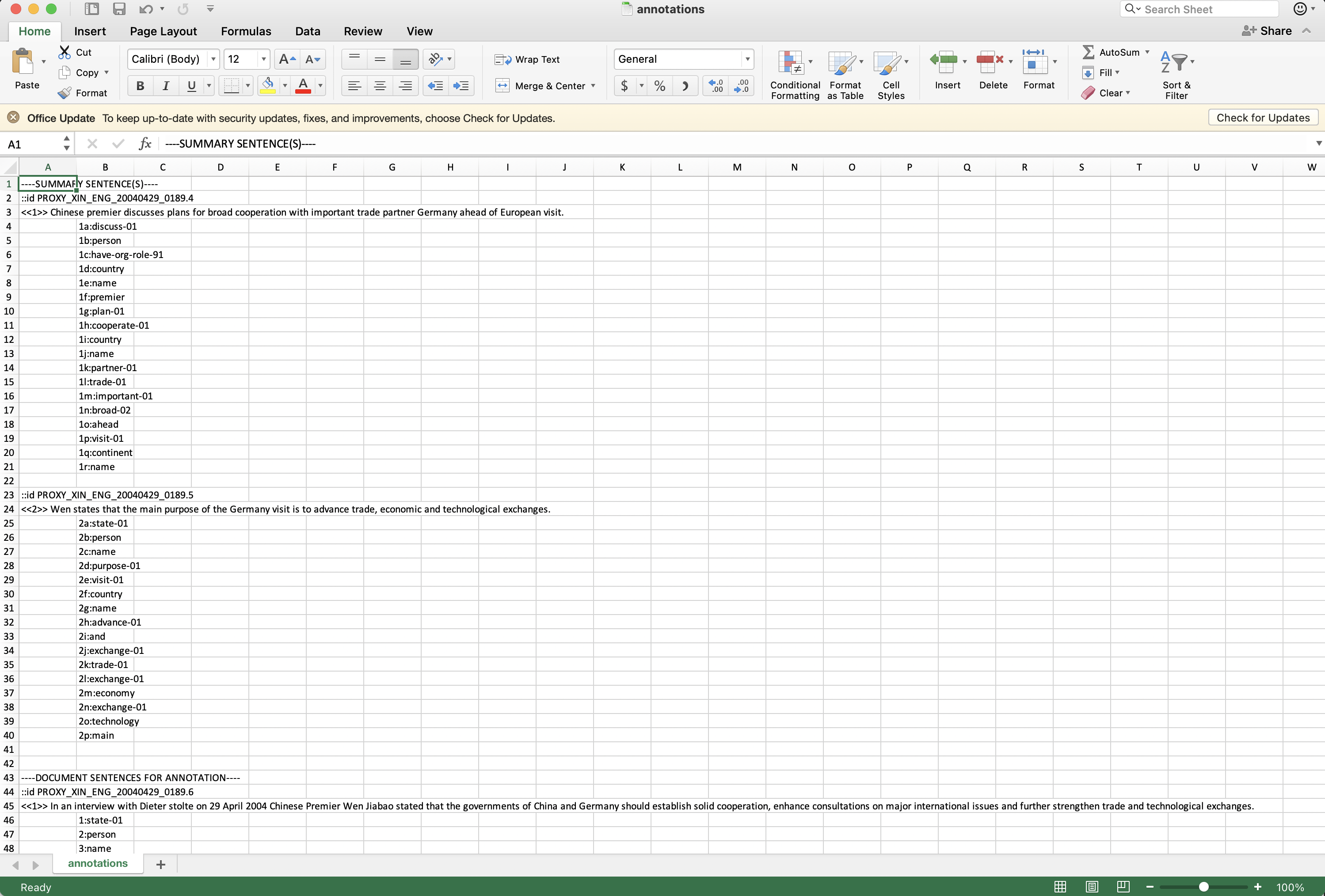}
    \caption{An example of the annotation interface for document-summary alignment.}
    \label{fig:spreadsheet}
\end{figure*}

The main annotation interface is a spreadsheet containing the minimal information for summary and document sentences. An example is provided in \autoref{fig:spreadsheet}.

%\kmnote{This sounds fishy. How do you know when they will need the visualiztion? Won't that bias them in doing the annotation? Could you instead be more specific about when this is needed? For example, could you have algorithmically decided when to provide the visualization based on the AMR structure? }
%\fnote{Whoops, I guess that was unclear - I meant that we provide the visualization for all sentences, just in case they need it.}
For many entities, the sentence text and concept label alone is sufficient to determine the role the node plays in the sentence; however, in case further disambiguation is necessary, we also provide both the AMR-to-text alignments in an HTML file for each document as well as a labeled visualization of each sentence graph.
%KM appendix fine. 
%\kmnote{Is there an interface? That sounds better than just using a spreadsheet.} \fnote{the interface was a spreadsheet... we can omit that detail and leave it for the appendix, I suppose...}  \nvnote{perhaps a screenshot would be helpful?} \fnote{I was going to include one in the appendix - do you think it would be better here?}

\begin{figure}[H]
    \centering
    \includegraphics[width=7cm]{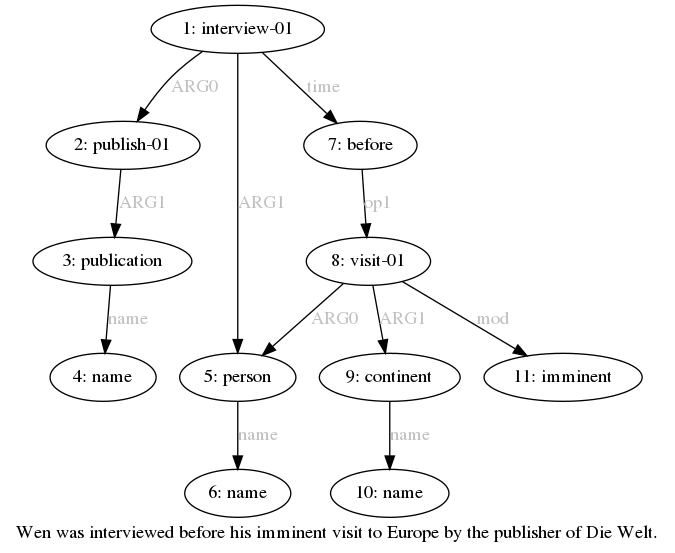}
    \caption{An example of the graph visualization.}
    \label{fig:graph}
\end{figure}

\begin{figure*}
    \centering
    \includegraphics[width=6in]{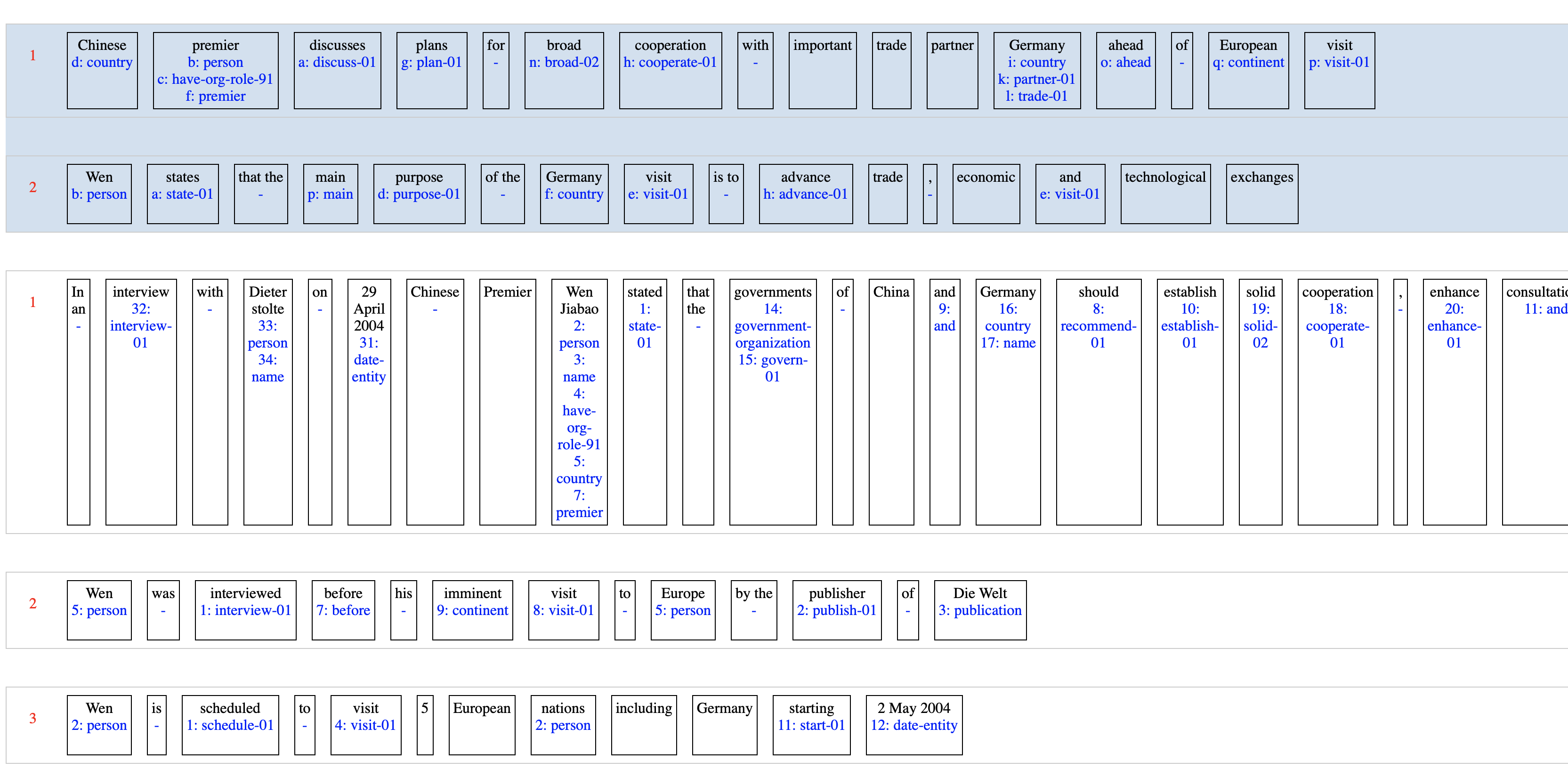}
    \caption{An example of the HTML node-to-text alignment visualization.}
    \label{fig:alignments}
\end{figure*}

Supplemental visualizations of the AMR graph structure and of the node-to-text alignments for each sentence are provided as separate files. Examples are provided in \autoref{fig:graph} and \autoref{fig:alignments}.

For each document node in the main spreadsheet, the annotators are asked to provide the following:
%KM not needed. Sounds informal.
%two things: 
(1) the space-separated list of handles for the set of summary nodes that refer to the same specific entity as that document node, if any (for example, ``1b 2f"), and (2) an additional tag marking the alignment list as ``abstractive" if the document and summary nodes refer to the same instance of an idea but are different concepts - for example, if a university ``organizes" a workshop in the document, but ``funds" it in the summary, those two verbs would be an abstractive match - or if one is an abstraction of the other in another way, e.g. one is an aggregate that includes the other. We do not consider the abstraction tags in this paper, but they would lend themselves to future work on more operations such as aggregation and sentence fusion.
\section{Types of errors in concept merging}
\label{sec:error-analysis}
\begin{table*}[t]
    \centering
    \begin{tabular}{c|p{10cm}}
        \toprule Error type & Example sentence fragment pair that produces error\\ \midrule
        \multirow{2}{*}{Stopword conflation} & ...a failure for Swedish law \textbf{and} order... \\
        & Both Norway \textbf{and} Sweden... \\ \hline
        \multirow{2}{*}{Missing synonyms} & ...nuclear \textbf{bombs}...\\
        & ...nuclear \textbf{arms}...\\ \hline
        \multirow{2}{*}{Name skipping} & Estonian informatics center spokeswoman \textbf{Katrin pargmae} stated...\\
        & \textbf{Katrin pargmae} also stated...\\ \hline
        \multirow{2}{*}{Concept conflation (surface)} & ...required to inform any person whose \textbf{assets} are being frozen... \\
        & ...an European Union decision to freeze the \textbf{assets} of the People's Mujahadeen of Iran.\\ \hline
        \multirow{2}{*}{Concept conflation (hidden)} & ...the foreign \textbf{minister} of India... \\
        & ...the foreign \textbf{minister} of China...\\\bottomrule
    \end{tabular}
    \caption{Examples of common error types in concept merging.}
    \label{tab:errors}
\end{table*}

The first two types of errors are relatively easy to fix. Stopword conflation occurs whenever there are multiple instances of a stopword present in the document,
%(most frequently ``and" or the organizational membership concept ``have-org-role-91", which typically appears in the text as ``of", e.g. as in ``the foreign minister of India")
and can be easily avoided by simply holding out stopword nodes from the concept merge process. Nodes that have synonymous but not identical labels will not be merged even when they refer to the same entity; % such as in the third example in \autoref{tab:errors}: ``bombs" and ``arms" are not considered the same concept by concept merging.
this can be approximately solved by using a high word embedding similarity threshold rather than exact concept matching.

The third error, name skipping, is a product of concept merging's initial name and date collapse step, in which name and date nodes are merged with their parents if they are only children, renaming the parent nodes with the combined concept. Name skipping occurs when a name node is merged into its parent in some instances but not in others, which splits the resulting merge clusters into those that were identified by name and those that were not. In the second example in \autoref{tab:errors}, the first instance of ``Katrin pargmae" will not be merged with its name node as it has additional children (the organization role of spokeswoman), but the second instance will. This is an inherent issue of the name collapse step, and can be addressed by replacing it with a more robust method of detecting co-reference.

The final two types of errors can be viewed as special cases of the same error, namely, merging entities that are different instances of the same entity type, but the method of addressing the error differs between cases. In the first case, a node representing an entity that is directly mentioned in the text is erroneously merged with another of the same type. To address this, we would need to perform text co-reference to determine whether nodes of the same type directly co-refer.

%In the second case, nodes that are not explicitly mentioned in the text but appear in the AMR as implicit attachments to an entity that is explicitly mentioned, and are erroneously merged with other implicit nodes of the same type.
%KMTues - The above sentence has no main verb that I could find. I've edited to make less complex. You may want to work on it to improve fluency more. 
In the second case, nodes that are not explicitly mentioned in the text, but instead appear in the AMR as attachments to a node that does explicitly appear, are erroneously merged with other implicit nodes of the same concept type. Concept merging cannot distinguish between the hidden person nodes in this case, even though they represent entities that do not co-refer.

It is this final type of error that the person-focused and combined merge strategies we present are designed to directly address: the descendant subtree of any person-type entity mentioned in the text must be merged only with the subtrees of other person nodes coreferent with that entity. We focus on person concepts specifically as 
%we note that 
this is a very common case that often induces a large subtree. % EMNLP
\section{Content Selection Model}
\label{app:content}
As our classification model we use a feedforward layer attached to a Graph Attention Network, or GAT \cite{velickovic2018graph}, that uses a modified the scaled dot-product attention function of the Transformer architecture \cite{vaswani-etal-attention} that is masked according to the undirected AMR graph structure. We use two layers, a single attention head, 256-dimensional intermediate representations, and the ReLU activation function. We implement our model in PyTorch. Node parameters in the GAT layers are initialized from a unit normal distribution; the classification layer is initialized with the PyTorch default. % FL: ...I actually don't know what this is. oh well
% in the future, i have found that using an initializer is super helpful! you live you learn you retrain 
% FL: yeah it definitely helped a lot for the GAT, i don't know why i didn't think to try it for the feedforward layer. ah well. next time
% FL: live, learn, get a restraining order against neural nets,

The input representations for each node consist of the summed word embeddings for all words in its concept label, as well as four discrete features: in-degree, out-degree, sentence index and number of occurrences in the text. The representation of every node at each graph layer is recalculated as an attention-weighted sum over the representations of itself and its neighbors in the AMR graph. The final node representation is passed to the single-layer feedforward classification component.

We train our classifier with the Adam optimizer with a learning rate of .001 and batch size of 1 for a maximum of 128 epochs, using an early stopping routine that halts training after training loss decreases by any amount between epochs or development loss has not increased within the last three epochs.

Using Transformer-style self-attention (masked to respect the AMR graph), the learnable parameters for the GAT are a projection matrix for keys, queries and values for each layer. As we use two layers, an intermediate representation size of 256, and 304-dimensional inputs, the total number of parameters for the GAT is therefore $2 \times (256 \times 304 + 304 \times 256 + 304 \times 304)$. The binary classification layer has $304 \times 2$ parameters.

A single training run of this model over the full proxy report dataset typically runs for around 30 epochs (with early stopping) and takes approximately three minutes on a single Tesla V100 GPU.
\section{Combined Merge Pseudo-code}
\label{app:person}
We present pseudocode for the full combined merging strategy in Algorithm \ref{alg:merge}.

\begin{algorithm}
\SetAlgoLined
\KwIn{A collection of sentence graphs $D = \{S_1, S_2, ..., S_n\}$, where $S_i = (V_i, E_i)$.}
// Initialize document graph with artificial root\\
$V \gets \bigcup_{i=1}^n V_i \cup \{-1\}$\\
$E \gets \bigcup_{i=1}^n E_i \cup \{(-1, root(S_i)) \mid i \in [n]\}$\\
// Collect subgraphs and aligned text spans\\
p $\gets \{n \in V_i \mid label(n) = person\}$\\
c $\gets \{descendants(n) \mid n \in$ p$\}$\\
// Assign cluster ids to subgraphs via spans\\
spans $\gets \{\{\bigcup span(n) \mid n \in C_i\} \mid C_i \in $ c$\}$\\
ids $\gets \{cluster\_id(span(p_i))$ if $p_i$ is clustered else $cluster\_id(C_i) \mid C_i \in$ c$\}$\\
// Partition coreferent type matches\\
$P \gets \{\}$\\
\For{$k \in cluster\_ids$}{
    cluster $\gets \{C_i \in$ c $\mid ids[i] = k\}$\\
    $C \gets \{n \in C_i \mid C_i \in$ cluster$\}$\\
    \For{unique $l \in \{label(n) \mid n \in C\}$}{
        $P \gets P \cup \{\{n \in C \mid label(n) = l\}\}$
    }
}
// Concept label partition on unmerged nodes\\
$U \gets \{n \in V \mid n \notin \bigcup_i \{P_i \in P\}\}$\\
\For{unique $l \in \{label(n) \mid n \in U\}$}{
    $P \gets P \cup \{\{n \in U \mid label(n) = l\}\}$
}
// Merge from partition\\
$V \gets \{i | P_i \in P\}$\\
$E \gets \{(i, j) | P_i, P_j \in P, \exists k \in P_i, l \in P_j \ni (k, l) \in E\}$\\
\Return{$(V, E)$}\\
\caption{Combined merging}
\label{alg:merge}
\end{algorithm}

\section{Finetuning setup for BART generator}
\label{app:finetuning}
%\fnote{added for EMNLP.}

For each merge strategy, we start with a vanilla pretrained BART-large model 
and finetune it for four epochs %\fnote{specify other hyperparameters?}
with a learning rate of .00003
on the generation task where the input is linearized selected AMR and the output is the summary text. (We originally considered a range of up to 64 epochs, but found that validation performance according to automatic metrics %\fnote{specify?}
peaked before epoch 10 in all cases, usually between epochs 3-5.) %\cknote{For the future you would typically do early stopping per model on the validation set in case one model benefits from slighty more or less epochs. I think this is ok for now as the summary generation is not the main focus.}

The linearized AMR input is generated as follows: given a merged document graph and the output of a node selection model on that graph, i.e., a binary label for each node of the graph indicating whether that node is to be included in the summary or not, we linearize each sentence in sequence, keeping track of all nodes that have been seen before to avoid repetition. To linearize each sentence, we start at the original root node of that sentence (i.e., one of the children of the artificial root node of the document graph) and perform a depth-first traversal from that node. Nodes that we touch that are labeled 1 are added to the sequence; nodes that we touch that are labeled 0 are not added to the sequence unless they have a descendant labeled 1. Nodes that we have already touched are added to the sequence again, but their subtrees are not traversed again.
%\fnote{do i need to write up another algorithm for this?} \cknote{I think what you have here is good but if there's time add an algo -- definitely for the final draft.}
The final sequence is converted to a string in PENMAN format.

We present the pseudocode for the linearization procedure in Algorithm \ref{alg:linearize}.

\begin{algorithm}
\SetAlgoLined
\KwIn{A merged document graph $G = (V, E)$; labels $L$ for every node $n \in V$, with $L(n) \in \{0, 1\}$; original sentence roots $r_1, r_2, ..., r_s \in V$.}
// Initialize sequence\\
seq $\gets []$\\
touched $\gets []$\\
\For{$i \in [s]$}{
    // Traverse subtree of sentence root in depth-first order\\
    \For{$n \in$ descendant subtree of $r_i$}{
        \uIf{n $\in$ touched}{
            append $n$ to seq\\
            skip recursion and return to parent
        }
        \uElseIf{L(n) = 1 or a descendant of n is labeled 1}{
            append $n$ to seq\\
            continue recursion in depth-first order
        }
        \uElse{
            skip recursion and return to parent
        }
    }
}
\Return{seq}\\
\caption{AMR linearization}
\label{alg:linearize}
\end{algorithm}

%\section{Sample generated summaries}
%\label{app:generation}
%\fnote{maybe i should merge this with the finetuning section?? it feels like we have a second paper on generation in the appendix at this point} \cknote{:) Yeah I think you could merge for sure.}

%KM shoudl elaborate on this. I am. 
%Although it is not the focus of our work, we generate text summaries from the subgraphs selected in the node selection phase with combined merging.

For each merge strategy, we trained BART on linearized AMR for its selected subgraphs and the associated text summaries, and used this model to generate a text summary for each document. We provide examples of a few such summaries using the combined merge strategy in \autoref{tab:generated}.

\begin{table*}[]
    \centering
    \begin{tabular}{p{7cm}|p{7cm}}
        \toprule
        Gold summary & Generated summary output\\
        \midrule
        Afghanistan's counter-narcotic police have confiscated 1.2 metric tons (1.3 tons) of opium during a raid. In 2007 Afghanistan accounted for 93 percent of the world's opium supply. & Afghanistan's Interior Ministry announced the confiscation of more than a ton of opium from a drug trafficker and the detention of a drug smuggler in the southernmost province of Helmand. The drug was smuggled into Afghanistan from Pakistan.\\ \hline
        Russia proposes cooperation with India and China to increase security around Afghanistan to block drug supplies. Afghanistan is 1 of the world's major opium producers and supplies Western markets through transit countries such as Russia. & Russia's Foreign Minister Lavrov stated that the Russian government will increase security at the borders with Afghanistan and Tajikistan in order to decrease the flow of illegal drugs from Afghanistan to the world market. The Russian government stated that Afghanistan is the world's largest producer of opium and the largest supplier of heroin to Russia.\\ \hline
        Russian Federation President Dmitry Medvedev promised on May 15, 2008 to provide funding for Russia's nuclear missile program. & Russian President Dmitry Medvedev made his first public appearance since being sworn in as the new President of the Russian Federation on November 9, 2008. The Russian President stated that the Russian military is ready to respond to any threat from the West.\\ \hline
        \bottomrule
    \end{tabular}
    \caption{Sample generated summary output from BART given our selected content as input.}
    \label{tab:generated}
\end{table*} % EMNLP
\section{Human evaluation of generated summaries}
\label{app:humaneval}
%\fnote{added for EMNLP.}

\begin{figure*}
    \centering
    \includegraphics[width=6in]{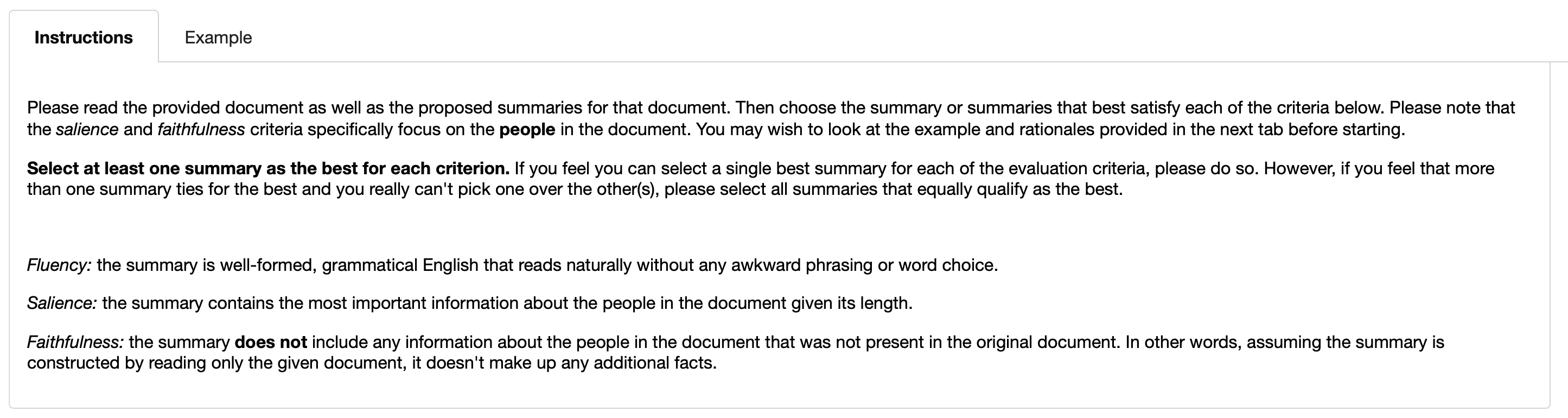}
    \caption{Instructions for human evaluation.}
    \label{fig:eval-instructions}
\end{figure*}

\begin{figure*}
    \centering
    \includegraphics[width=6in]{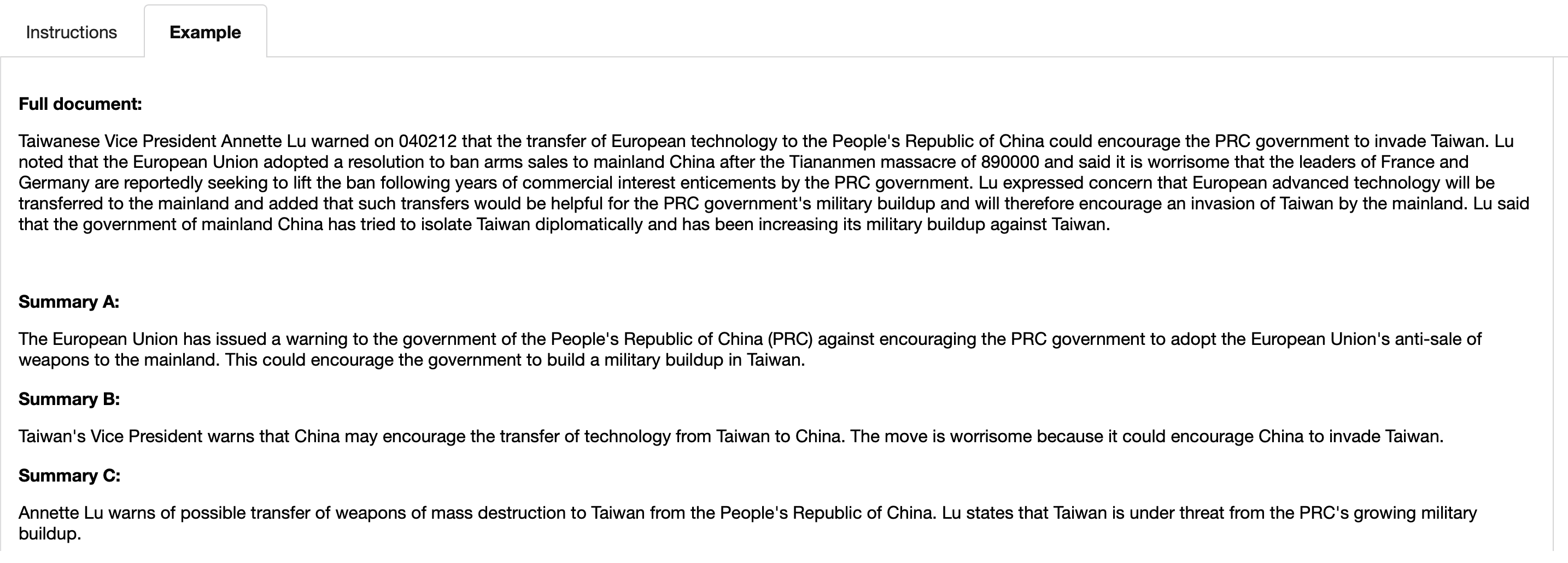}
    \includegraphics[width=6in]{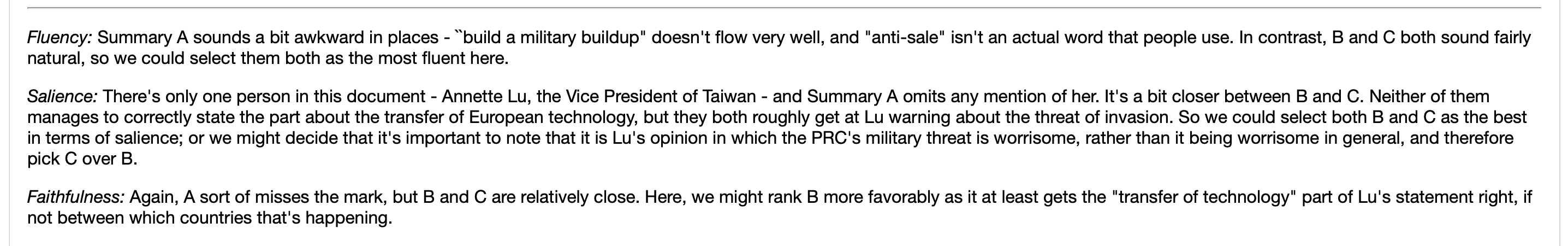}
    \caption{Reference example given for evaluation.}
    \label{fig:eval-example}
\end{figure*}

\begin{figure*}
    \centering
    \includegraphics[width=6in]{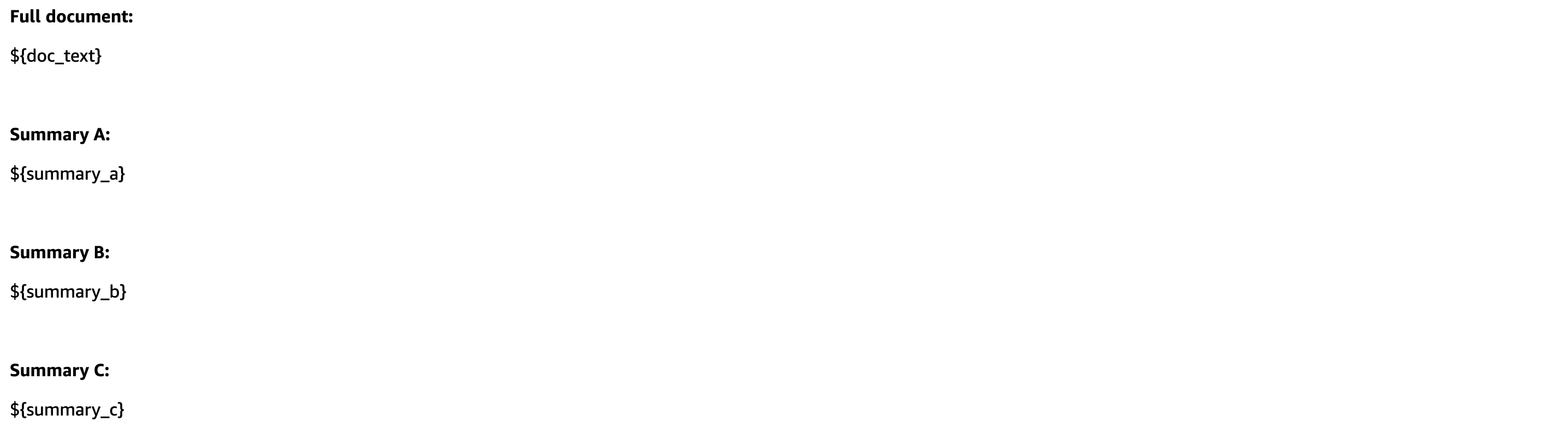}
    \includegraphics[width=6in]{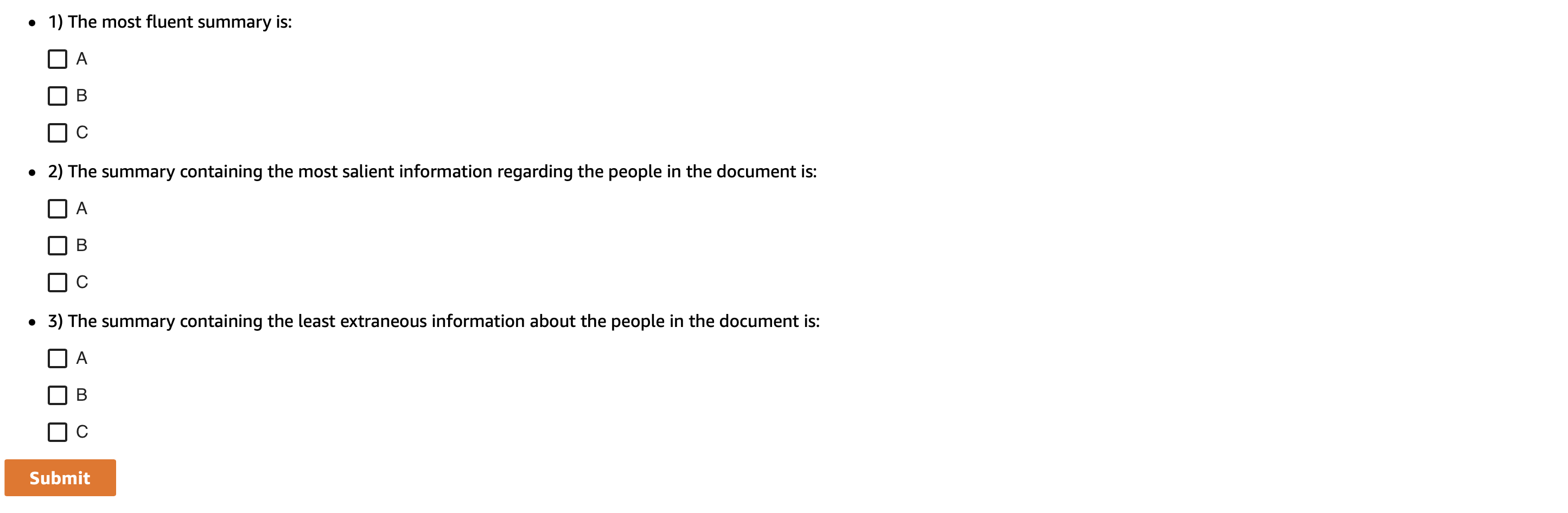}
    \caption{Document and summary text placeholders and ranking prompts for evaluation.}
    \label{fig:eval-task}
\end{figure*}

We recruited five annotators with a background in NLP to perform an evaluation of the summaries generated with the three different merge strategies. Of the 33 test set documents, each was assigned two annotators. For each document, annotators were given the first ten sentences of the document text as well as the three summaries produced on that text by the different merge strategies, and asked to choose the best summary among those on each of three criteria: fluency, salience and faithfulness, specifically with respect to people mentioned in the document. The specific annotation instructions, reference example, and task layout are provided in Figures \ref{fig:eval-instructions}, \ref{fig:eval-example}, and \ref{fig:eval-task}, respectively.

 % EMNLP

\end{document}